\documentclass[conference]{IEEEtran}
\usepackage{cite}
\usepackage{url}
\usepackage{amsmath,amssymb,amsfonts}
\usepackage{algorithmic}
\usepackage{subcaption}
\usepackage{graphicx}
\usepackage{listings}

\lstdefinestyle{mystyle}{
    basicstyle=\ttfamily\small,
    breaklines=true,
    frame=single,
    xleftmargin=10pt,
}

\usepackage{textcomp}

\usepackage{xcolor}
\usepackage{comment}
\def\BibTeX{{\rm B\kern-.05em{\sc i\kern-.025em b}\kern-.08em
    T\kern-.1667em\lower.7ex\hbox{E}\kern-.125emX}}
\begin{document}

\title{Precise and Efficient Orbit Prediction in LEO with Machine Learning using Exogenous Variables \\
}


\author{\IEEEauthorblockN{1\textsuperscript{st} Francisco Caldas}
\IEEEauthorblockA{\textit{NOVA LINCS} \\
\textit{NOVA School of Science and Technology}\\
Caparica, Portugal \\
f.caldas@campus.fct.unl.pt}
\and
\IEEEauthorblockN{2\textsuperscript{nd} Cláudia Soares}
\IEEEauthorblockA{\textit{NOVA LINCS} \\
\textit{NOVA School of Science and Technology}\\
Caparica, Portugal \\
claudia.soares@fct.unl.pt}

}

\maketitle

\begin{abstract}
The increasing volume of space objects in Earth's orbit presents a significant challenge for Space Situational Awareness (SSA). And in particular, accurate orbit prediction is crucial to anticipate the position and velocity of space objects, for collision avoidance and space debris mitigation. When performing Orbit Prediction (OP), it is necessary to consider the impact of non-conservative forces, such as atmospheric drag and gravitational perturbations, that contribute to uncertainty around the future position of spacecraft and space debris alike.
Conventional propagator methods like the SGP4 inadequately account for these forces, while numerical propagators are able to model the forces at a high computational cost.
To address these limitations, we propose an orbit prediction algorithm utilizing machine learning. This algorithm forecasts state vectors on a spacecraft using past positions and environmental variables like atmospheric density from external sources. The orbital data used in the paper is gathered from precision ephemeris data from the International Laser Ranging Service (ILRS), for the period of almost a year. We show how the use of machine learning and time-series techniques can produce low positioning errors at a very low computational cost, thus significantly improving SSA capabilities by providing faster and reliable orbit determination for an ever increasing number of space objects.

\end{abstract}


\section{Introduction}

Since the late 1950s, when the first artificial satellite was launched, the number of resident space objects (RSOs) has steadily increased. Presently, Earth's orbit is host to an estimated one million objects larger than $1$ cm, with only $35{,}000$ objects exceeding $10$ cm regularly tracked~\cite{ESA2023}. To add to this, an additional $10{,}000$ satellites are expected to be launched by the year 2030, as per existing estimates and licensing arrangements~\cite{Lawrence2022}. To protect the space environment in Low-Earth Orbit, and due to the compounding effects of space pollution that can create a chain reaction of collisions, known as Kessler Syndrome, it is indispensable to accurately track and predict space debris and satellites’ orbits. The predominant methods to predict the future position and velocity of a satellite (state vectors) are physics-based methods that can be divided into analytical and numerical approaches. Numerical methods obtain the future state of the space objects by integrating the equations of motion of the satellite or debris, and taking into consideration the conservative and non-conservative forces applied on the space object. These methods are highly precise, at the expense of being computationally costly. Common propagators include Dormand-Prince 8(7) (RKDP8) and Runge-Kutta-Nystrom 12(10) (RKN12)~\cite{Montenbruck2000,Jones2012} and predictor-corrector methods, namely Adams-Bashforth-Moulton (ABM) and Gauss-Jackson (GJ)~\cite{Poore2016}.
Analytical methods leverage closed-form equations derived from simplified models of orbital dynamics, offering expedited computations. However, such approaches, exemplified by the well-known Simplified General Perturbations 4 (SGP4) method~\cite{Vallado2006}, exhibit limitations in long-term orbit predictions due to their reliance on overly simplified force models. 

Orbit Propagation is a key component in any space surveillance framework, as it allows to determine possible close approaches between space objects in the future. When two space objects are in a close approach, this is called a conjunction, and, as the number of space objects increases, so the number of conjunctions is to increase exponentially. According to the modelling presented in Long \cite{long2020}, a target population of $10{,}000$ active CubeSat satellites positioned at $600$ km is projected to experience over $350$ collisions during $30$ years. Consequently, the development of an algorithm that can rapidly and precisely predict the propagation of an increasing number of satellites and debris is paramount.


\subsection{Related Work}

To bridge the gap between time-consuming numerical methods and inaccurate analytical models, error-correction techniques have been applied in a series of papers~\cite{Sang2017,Bennett2012,SANJUAN2017254}, all using a similar approach, that involves correcting the SGP4 model using historical data specific to a given satellite with an error correction function \cite{CALDAS202497}. The same methodology was employed using different machine learning algorithms: Support Vector Machines (SVM), Artificial Neural Networks (ANN) and Gaussian Processes (GP)~\cite{Peng2019,Peng2018a,Peng2019a}, in a series of papers by Peng and Bai~\cite{Peng2019}. The authors introduced three types of ML generalizations for Orbit Prediction. Type-I generalization involves interpolation, accurately determining an object's state for an unknown section based on initial and final states. Type-II, forecasting, trains an algorithm to predict the future position of a space object beyond the training data timeframe. Type-III extends ML capabilities to extrapolate information from one object to predict the orbit of a different space object \cite{Peng2019}.

Under the same hybrid ML-SGP4 error-correction approach, five independent works show that different machine learning architectures: Convolution Neural Networks (CNN)~\cite{Pihlajasalo2018}, Artificial Neural Networks (ANN)~\cite{San-Juan2018}, Recurrent Neural Networks (RNN)~\cite{Curzi2022}, Gradient Boosting Trees (GBT)~\cite{Li2021}~\cite{Li2020}, and Long Short-term Time-series network (LSTMNet)~\cite{Hong2023}) can be used to improve the SGP4 Orbit Prediction. A step forward has been proposed using a differentiable SGP4 ($\delta$SGP4) that is optimized as a layer within a multi-layered neural network \cite{acciarini}.

A different approach, without using the SGP4 model, has also been experimented with:
In Muldoon et al~\cite{Muldoon2009}, the SGP4 model was approximated using a polynomial fit. In 2012, a different method was introduced using Latent Force Models (LFMs)~\cite{Hartikainen2012} to combine a numerical integrator with non-parametric data-driven components. This work was extended by Rautalin et al \cite{Rautalin2018}, who obtained favorable results for a set of satellite constellations in Medium Earth Orbit (MEO) and
Geo-Synchronous Orbit (GEO). 

\subsection{Contributions}

This paper proposes to expand the state-of-the-art approach by performing orbit determination without using a hybrid error-correction approach. Any error-correction approach is, by definition, more computationally costly than the original model it is trying to correct, and in this case, there is also the constraint of using the unique input necessary for the SGP4 model. Our approach is not bounded by these limitations, and can further reduce computation time in a real-time setting. By using exogenous variables relevant to obtain the effects of the non-conservative forces, we are replicating the information available to a Numerical Propagator, and training and testing our results in a real-life dataset, that necessarily represents more accurately the complex reality of orbit prediction.

To summarize, this paper presents:

\begin{itemize}
    \item An advancement in orbit prediction by eliminating the reliance on a hybrid error-correction method;
    \item The integration of exogenous variables to capture the effects of non-conservative forces;
    \item The testing in a real world setting with a data-centric approach;
    \item A machine learning model outperforming in error a J2 numerical propagator, using less compute time;
    \item Empirical validation of computational costs.

\end{itemize}

\section{Background}

\subsection{Orbit Dynamics}

The complete equation of motion of a satellite in a given coordinate system~\cite{Horwood2011,Vallado2001} can be written in the form
\begin{equation}
\label{sample:2}
f(\boldsymbol{x},t) = \ddot{\boldsymbol{r}} = - \frac{\mu_\oplus}{r^3}\boldsymbol{r} + \boldsymbol{a}_{pert}(\boldsymbol{r},\dot{\boldsymbol{r}},t) ,
\end{equation}
where $r = \|\boldsymbol{r}\|$ is the norm of the position vector, $[\boldsymbol{r} \quad \dot{\boldsymbol{r}}]^T= \boldsymbol{x} $, and $\mu_\oplus$ is the gravitational constant multiplied by the masses of the Earth and the RSO, with the perturbing forces being
\begin{equation}
    \boldsymbol{a}_{pert}(\boldsymbol{r},\dot{\boldsymbol{r}},t) = \boldsymbol{a}_{NS} + \boldsymbol{a}_{NB} + \boldsymbol{a}_{drag} + \boldsymbol{a}_{SRP} + \boldsymbol{a}_{tides}  + \boldsymbol{a}_{others} ,
\end{equation}
where each term is, in order, the perturbation of Earth's gravity potential, due to a Non-Spherical Earth ($\boldsymbol{a}_{NS}$), the N-body perturbations of the Sun, the moon and other planets ($\boldsymbol{a}_{NB}$), the atmospheric drag ($\boldsymbol{a}_{drag}$), the  Solar Radiation Pressure ($\boldsymbol{a}_{SRP}$), tidal effects ($\textbf{a}_{tides}$) and others perturbations ($\boldsymbol{a}_{others}$) that include General Relativity theory adjustments and geomagnetic pulls.

For a given initial condition $\boldsymbol{x}(t_0) = \boldsymbol{x}_0$, the position of a satellite at any point in time can be implicitly written as the solution of the equation flow $\phi$:
\begin{equation}
\boldsymbol{x}(t) = \phi(t;\boldsymbol{x}_0,t_0).
\label{equation:flow}
\end{equation}
where $\phi(t;\boldsymbol{x}_0,t_0)$ can be approximately solved using a numerical solution or any analytical approximation. In the case of this paper, $\phi$ will be approximated using a Neural Network architecture, with the initial condition expanded to a set of previous states and other features.

\subsection{Dataset}

The main data used in this work is comprised of Consolidated Prediction Format (CPF) files, that provide very precise ephemerides files for a series of satellites. In this case we choose the LARETS satellite, and data ranging from 2022-01-01 to 2023-01-01. Each file contains $\left(\boldsymbol{r} =\left(x,y,z \right),t \right)$, position and time, at $3$ minutes intervals, for a duration of $5$ days after the file's production date. 
To establish ground truth data, we adopt the assumption that predictions with shorter forecasting periods have greater accuracy. Thus, we construct a single time-series of positions by selecting the prediction closest to the production date for each time step. For instance, if multiple CPF files contain information for a given time step, we prioritize the most recently produced file.

\begin{table}[b]
\caption{Summary of the exogenous features}
\begin{center}
\begin{tabular}{|c|c|}
\hline
\textbf{\# of Features}&\textbf{Exogenous Features} \\
\hline
\hline
\textbf{8} & Space Weather Drivers  \\

\textbf{3} & Velocities \\

\textbf{6+6}& Keplerian and Equinoctial\\

\textbf{1} & Total Mass Density \\

\textbf{3} & Gravity field Functionals\\

\hline
\end{tabular}
\label{tab2}
\end{center}
\end{table}

The data was divided into training , validation and test, with the testing set($15\%$) representing roughly the last two months of the dataset. When performing evaluation of ML time-series models, it is particularly important to be mindful of possible time leakages. In any splitting scheme, the training data must always be from a time period before the validation or testing set.

The selection of exogenous variables, summarized in Table \ref{tab2}, was guided by their relevance in modeling the non-conservative forces acting on the satellite \cite{gondelach2020a}. Subsequently, these variables underwent pruning to eliminate highly correlated or unnecessary features. 

Space weather drivers are commonly used as input to air density models, and therefore are relevant as indicators of space weather conditions and also as proxies for density estimation. From the {\small{\textsc{High Resolution OMNI Data (Omniweb)}}} website eight features pertaining to space weather were extracted, namely: Interplanetary Magnetic Field (IMF) magnitude \cite{Rastogi1975}, Solar Wind Plasma wind speed \cite{solarwind}, temperature and density \cite{solarwind2}, and derived parameters such as Flow Pressure, Electric Field and Alfven Mach Number \cite{Nishino2022}. Finally two hourly indices, the kp-index \cite{Kp_Index} and the Lyman-alpha index \cite{milligan2020}. 

Additionally, Keplerian $(a, e, i, \omega, \Omega, \nu)$ and modified equinoctial $(p, f, g, h, k, l)$ coordinates were taken into account due to their demonstrated potential for improving orbit propagation and uncertainty estimation by minimizing non-linearities  \cite{Aristoff2021}\cite{Bau2021}. The derivation of velocities in Cartesian coordinates relied on a Hermite interpolation filter, pivotal for obtaining these coordinates.

In this study, the NRLMSISE‐00 \cite{picone} model was employed to derive the Total Mass Density specific to the satellite state at each time. Similarly, the GOCO06s \cite{kvas2019goco06s} gravity field model was used to obtain the gravity perturbation affecting the satellite at each time-step.

\section{Network Architecture}

The problem is approached as a multi-step forecasting task, wherein it is customary to utilize $m$ states preceding the forecasting start time ($t_0$) to predict all future steps at once. This stands in contrast to an Auto-Regressive forecasting framework, which predicts only one step ($t+1$), $n$ times.

\begin{figure}[b!]
     \centering
    \includegraphics[width=1.\columnwidth]{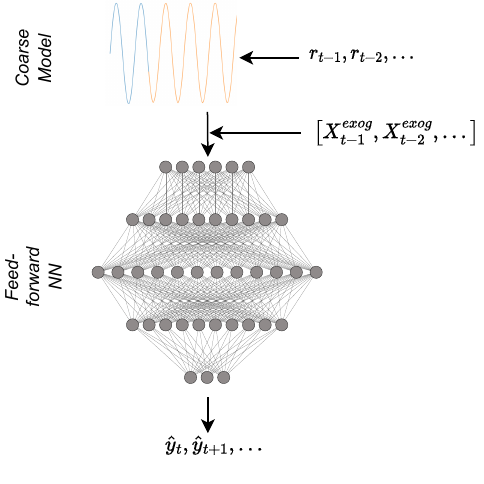}
     \caption{Two-layer Model Architecture: Integration of two independently trained models, where one serves partially as the input source for the other.}
     \label{fig:model}
\end{figure}

The overarching architecture employs a two-step structure. The initial step, referred to as the coarse model, is designed to provide an initial prediction. Subsequently, a second step, independently trained, refines the initial prediction by incorporating covariates, as can be seen in Figure \ref{fig:model}. 
This two-step model is used to ideally separate the main effect of the two-body dynamics, with the effect of the perturbation forces.

\subsection{Coarse Model}

The inaugural stage of this architecture is characterized by the implementation of a ``coarse model''.
The Coarse Model assumes two critical roles. Firstly, it performs a strategic reduction in the search space that the Feed-Forward NN, the second step of the model, must navigate, and thus reducing the range of potential solutions. 
Secondly, it provides an initial approximation of the satellite's orbital state for the forecasting period. In this study, we propose two distinct approaches for the coarse model. The first is a robust and straightforward statistical time-series model, while the second is a highly complex neural network tailored for time-series problems. Each approach has its own set of advantages and drawbacks.

\subsubsection{Prophet}

The Prophet model~\cite{Taylor2017} is an automated model built upon a decomposable additive model. Usually, Prophet encompasses both trend and seasonality components, along with the incorporation of point-wise effects. In this paper, the additive equation can be simply written as:

\begin{equation}
    \hat{x}_t = s_0(t) + s_1(t) + s_2(t) + s_3(t) +\epsilon
\end{equation}

where $s_i$ are the $4$ seasonal components of the time-series and $\epsilon$ is a white noise term. Unlike the usual time-series additive models, there is no trend component, and no event component. Due to the automated fitting process and its simplicity, this model is a very robust albeit limited model of the orbit dynamics. The robustness is rather important, as it guarantees that forecasting errors are constant across training and test set. However, the robustness comes at a computational cost, particularly during both the training and evaluation stages, as each Prophet model is trained on the preceding $1000$ time-steps for every sample. 

\subsubsection{iTransformer}

The Inverse Transformer (iTransformer) \cite{liu2023itransformer} is a Transformer model~\cite{Vaswani} modified for multivariate time-series. The Transformer architecture has shown state-of-the-art results in Natural Language Processing~\cite{DBLP:conf/naacl/DevlinCLT19} and in Computer Vision (ViT \cite{dosovitskiy2020vit}) but as yet to show the same type of performance in time-series forecasting, particularly long time-series forecasting where linear models still achieve State-of-the-art (SOTA) results \cite{zeng2023transformers}\cite{ni2024mixture-of-linear-experts}.  The Inverse Transformer inverts the temporal tokenization of the Transformer, and tokenizes the entire time-series covariate, giving each variate its own token. Otherwise, the iTransformer follows the same architecture of the original Transformer. Compared with the Prophet model, this model is far more powerful at forecasting on its own, however, it is also more prone to overfitting, and therefore, the training of the second model needs to be performed on a separate dataset. 

\begin{table}[t]
\caption{iTransformer Hyperparameters after hyperparameter tuning}
\begin{center}
\begin{tabular}{|c|c|}
\hline
        \multicolumn{2}{|c|}{\textbf{iTransformer}} \\
        \hline
        \hline
        \textbf{encoder-decoder size}            & 573                                                 \\
        \textbf{\# attention heads}            & 11                                                  \\
        \textbf{encoder layers}           & 4                                                   \\
        \textbf{decoder layers}           & 1                                                   \\
        \textbf{dropout}             & $4.12\cdot10^{-3}    $                    \\
        \textbf{Dense NN size}               & 4046                                                \\
        \textbf{Activation} & GeLU \\
        \hline
    \end{tabular}
\label{tab1}
\end{center}
\end{table}

\subsection{Exogenous Variables Model}

The Feed-Forward Neural Network (FNN) is one of the more widely used neural network architectures. While not specifically tailored to time-series data, this architecture can, given the necessary depth and width and activation functions~\cite{SHEN2022}, approximate any continuous function, according to the Universal Approximation Theorem~\cite{pinkus_1999}. 
While this theorem indicates that given the proper weights and activation functions, any function can be approximated, that does not mean that it is possible to obtain, through gradient descent, the optimal weights. Therefore, an appropriate training scheme is necessary to reach an acceptable local minima.  

The data is feed to the model such that each sample is (time-steps, features), with features being the previous states, the coarse model forecast, and the exogenous variables. During this process, some exogenous variables were experimentally found to have no predictive power, namely the space weather drivers. Due to the use of the atmospheric density as one of the features, other space weather features became redundant, as they are partly indicators of the same effect (air drag), but not specific to the position of satellite. 

The model architecture is a 3-layer FNN, with 100 units per layer, LeakyRelu \cite{maas2013rectifier} activation functions and a dropout of 0.01. The simplicity of the model itself comes from the coarse model search space reduction, and also, through experimentation, the authors found that more complex models did not show improved results.

\subsection{Numerical Propagator}

\begin{table}[!htb]
\begin{center}
\caption{Numerical Propagator and forces considered for Propagation.}
\label{tab:propa}

\begin{tabular}{|c|c|}

\hline
        \multicolumn{2}{|c|}{\textbf{Numerical Propagator}} \\
        \hline
        \hline
\textbf{Integration}        &  Dormand Prince 8(5,3)     \\ 
 \textbf{Gravity Model}             &  200x200 (GOCO06s)   \\ 
 \textbf{Third-Body}      & Sun and Moon                   \\
 \textbf{Tides}    & Nan \\
 \textbf{atmospheric Model} & NRLMSIS00 \\
 \textbf{Solar Radiation Press.} & Nan \\
\hline
\end{tabular}
\end{center}
\end{table}

The numerical propagators presented in this paper were constructed using the \textsc{Orekit} library. The force model and the integrator for the precise numerical propagator are described in Table \ref{tab:propa}. The mass, ($10$ kg) and cross-section area ($0.031 $ m$^2$) was obtained from the DISCOS platform \cite{Mclean17}. 

The second numerical propagation algorithm, refereed to as J2, is a numerical Propagator using the same integrator, but considering only the J2 perturbation. 

%




\section{Results}

To evaluate the results two common forecasting metrics are used, the Mean Absolute Error (MAE):
\begin{equation}
    \text{MAE} = \frac{1}{3} \sum_{k=1}^3 \frac{1}{n}\sum_{i=1}^n |y_i^k-\hat{y}_i^k|
\end{equation} 
and the Root Mean Squared Error (RMSE):
\begin{equation}
    \text{RMSE} = \frac{1}{3} \sum_{k=1}^3 \sqrt{\frac{1}{n}\sum_{i=1}^n (y_i^k-\hat{y}_i^k)^2}
\end{equation} 
where the forecasting error is calculated for the propagation window and averaged across the three positional dimensions.

\subsection{Error Analysis}
\begin{table}[b]
\caption{Results for a $3$ days propagation in the test set. \footnotesize{*Numerical propagators and CPF ephemeris from NASA GSFC SLR have a reduced sample size, with a single ephemeris file produced each day.}}
\label{tab:results}
\begin{center}

\begin{tabular}{|c|c|c|c|}
\hline
\textbf{Model}        &  \textbf{MAE}    &  \textbf{RMSE}  & \textbf{Exec. Time} \\ \hline \hline
 \textbf{}             &  \textit{[m]}  &  \textit{[m]} & \textit{[s]}\\ \hline
{Prophet + FNN}       &  2277.69    & 2921.60  &     $1.41 \pm 0.189 $           \\
{iTransformer + FNN}  & 3019.23& 3719.67 & $0.0184 \pm 2.34 \cdot 10^{-6}$ \\
{J2*}   & 3944.43   & 5344.58&       $ 9.46 \pm 0.13$\\
{Num. Prop.*}         &  456.08          & 616.1397  &   $ 196.22 \pm 3.66$     \\ 
{Ephem. file*} & 14.92 & 23.89 & --\\

\hline
\end{tabular}
\end{center}
\end{table}

\begin{figure}[b!]
     \centering
    \includegraphics[width=0.51\textwidth]{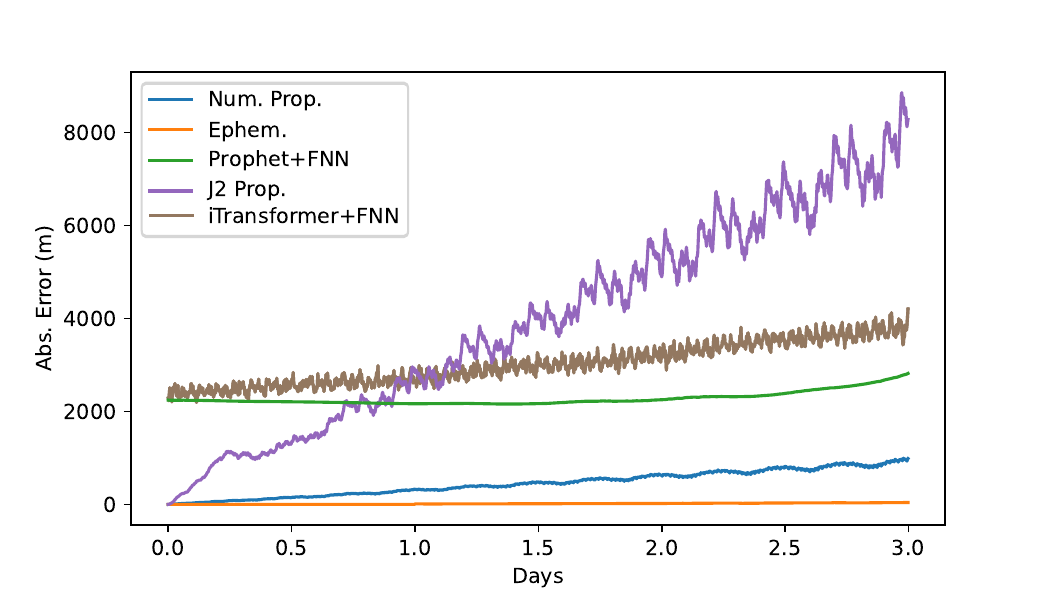}
     \caption{
The Mean Absolute Error (MAE) calculated over a 3-day propagation window reveals that the Numerical Propagation method excels in precision compared to the ML models. Nevertheless, as the propagation period extends, the error gap between the models diminishes. After approximately $1$ day, the ML models outperform the J2 model.}
     \label{fig:error}
\end{figure}
\begin{figure*}[h!]

     \begin{subfigure}[b]{0.49\textwidth}
         \centering
         \includegraphics[width=1.\textwidth]{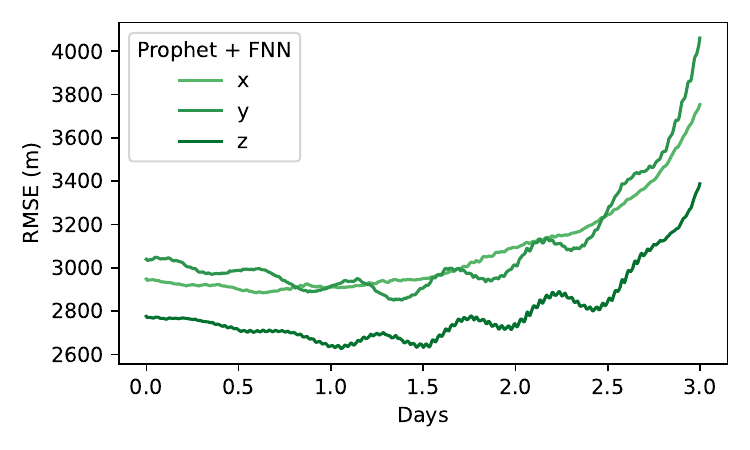}
         \label{fig:NNerror}
     \end{subfigure}
     \begin{subfigure}[b]{0.49\textwidth}
         \centering
         \includegraphics[width=1.\textwidth]{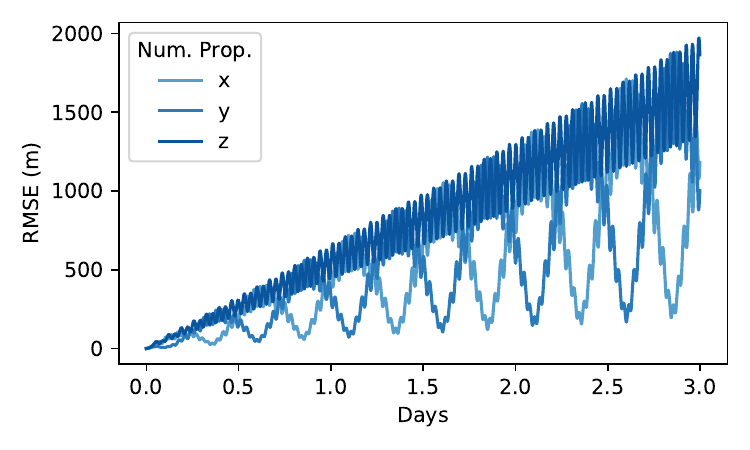}
         \label{fig:three sin x}
     \end{subfigure}
     
     \begin{subfigure}[b]{0.49\textwidth}
         \centering
         \includegraphics[width=1.\textwidth]{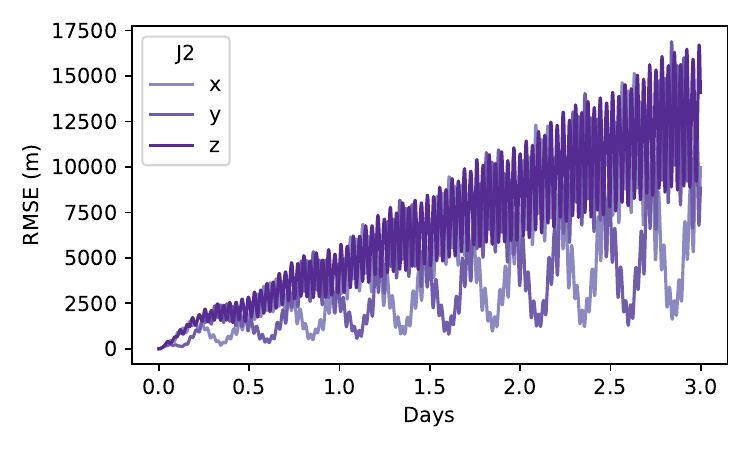}
         \label{fig:five over x}
     \end{subfigure}
     \begin{subfigure}[b]{0.49\textwidth}
         \centering
         \includegraphics[width=1.\textwidth]{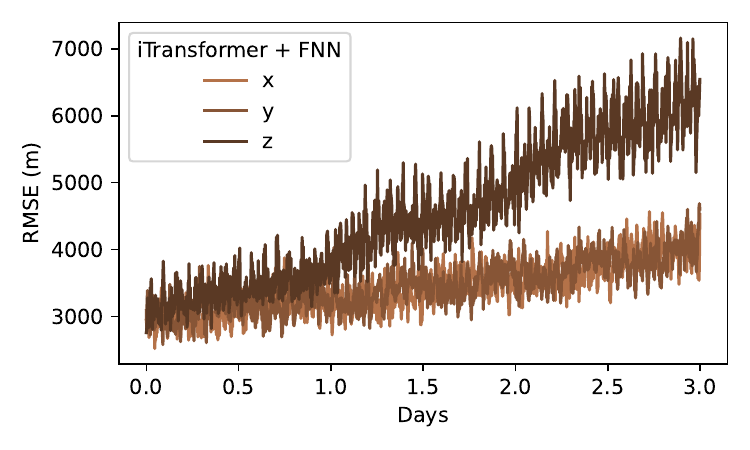}
         \label{fig:iTerror}
     \end{subfigure}
        \caption{RMSE in the $r = (x,y,z)$ Cartesian coordinates for the Prophet + FNN model, the Numerical Propagator, the  J2 Propagation and the iTransformer + FNN, respectively.}
        \label{fig:four graphs}
\end{figure*}
Table \ref{tab:results} shows that the average error for both ML models is higher than for the case of the numerical propagator. The clear advantage that a numerical propagator has over a completely empirical ML model shows how domain knowledge and having a proper force model are important. Nonetheless, the results of the model show some of the beneficial qualities that were to be expected, such as the mostly constant error across the three days forecasting window (three min. per step), particularly in the case of the Prophet + FNN model, as can be seen in Figure \ref{fig:error}. Furthermore, after one day, both models surpasses the J2 propagation, with the RMSE of the J2 being significantly higher for a three days propagation. Between both ML models, we observe that the simplest model (Prophet + FNN) presents better results, which is both indicative of the success of linear and seasonal models (such as the Prophet) in long-time forecasting, and also an indicator that further improvements might come from  physical knowledge injection instead of powerful NN architectures. The error difference between the accurate numerical propagator and the CPF files that were combined to determine the ground truth data indicate that the orbit propagation in these files is quite consistent across files, and that it is highly accurate, with a mean $14~m$ error over a three day period.

In Figure \ref{fig:four graphs} we can compare more in depth the orbit propagation error across the three days. In this Figure, we can observe that the $z$ coordinate is the most difficult to model, both for numerical propagators and the iTransformer+ FNN model, but is the one that has a lower error across time with the Prophet + FNN model. This might indicate that an additive seasonal model such the Prophet is particularly apt at modelling this variable. An alternative strategy could involve independently modeling each feature, taking advantage of the known benefits of smaller target sizes for learning. This approach may also open avenues for using alternative coordinate systems, such as equinoctial, directly as targets for the machine learning model.

\subsection{Computation Time}

Numerical Propagation is known to be computationally costly, and as the number of space objects in space increase exponentially over time, the bigger the burden is to concurrently propagate all these objects. To measure computation costs for the four models presented here, the authors used a Laptop with a Intel Core i7-10870H CPU with 16GB and a NVIDIA RTX 3060 8GB GPU. Models are coded in Python 3.8 and Pytorch 2.1.

\begin{figure}[h!]
     \centering
    \includegraphics[width=0.51\textwidth]{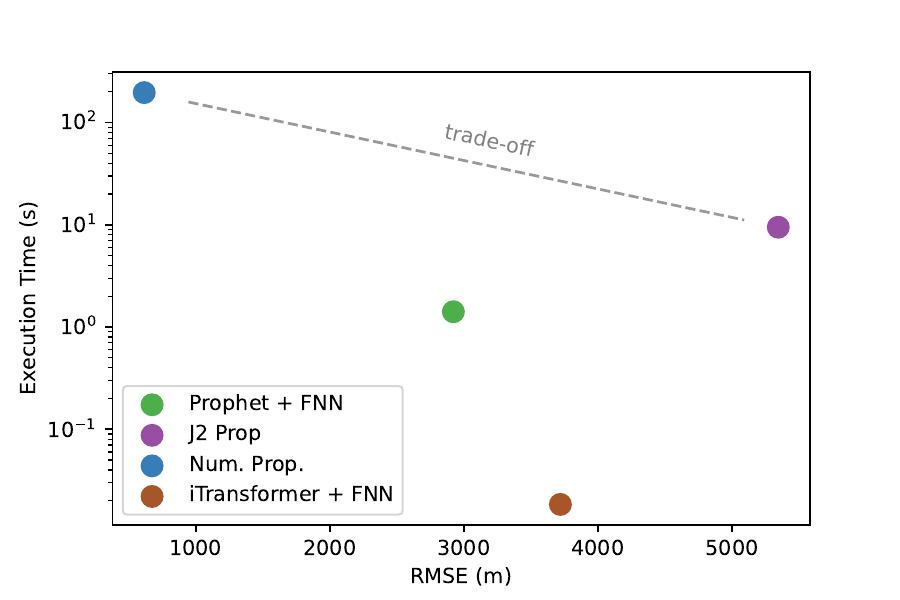}
     \caption{Average Execution time (in seconds) for a single propagation vs RMSE. Value obtained for data in the test set, across $\approx$ 2 months. The iTransformer + FNN model is the fastest model, propagating a state every $0.018$ seconds, at the expense of being less precise than the other ML model. The y axis is in logarithmic scale.}
     \label{fig:computation_time}
\end{figure}

In Figure \ref{fig:computation_time} we can observe the average computation time cost for a single propagation against its RMSE. 
The numerical propagator, a traditional approach, is found to be significantly more computationally expensive, with a cost exceeding 100 times that of the Prophet + FNN, and $10000$ times the iTransformer + FNN. In our experiments, the numerical propagator required an average execution time of $196.22 \pm 3.66$ seconds, while the complete iTransformer + FNN model, including data-preprocessing and feature augmentation through exogenous variables, exhibited a remarkably shorter execution time of $0.0184 \pm 2.34 \cdot 10^{-6}$ seconds. 

By decomposing the Prophet + FNN computational cost, it is revealed that the "coarse model", the Prophet time-series forecasting model, accounts for approximately $98\%$ of the overall execution time. Conversely, the pre-processing pipeline, responsible for integrating exogenous variables into the neural network (NN) model, contributes less than $1\%$ to the total computational cost. Notably, during the development of this work, this pipeline was highly optimized since the Prophet algorithm computation time is improved using previously trained parameters as a warm start. This highlights one of the biggest advantages of machine learning for orbit propagation, its computational velocity compared even with limited propagators, such as the J2 propagator. 

An element that is also present in Figure \ref{fig:computation_time}, where the y-axis is in logarithmic scale, is the numerical trade-off between RMSE and the computation cost. It is noteworthy that as RMSE decreases, the execution time experiences a substantial increase, for numerical propagator. This trade-off underscores the need for a thoughtful choice between model options. Specifically, when considering the RMSE, the ML models exhibit performance inferior to that of the numerical propagator. However, it is important to emphasize that the ML models offer a significant advantage in terms of computational efficiency, demanding substantially less execution time, 
In fact, the iTransformer + FNN is the most efficient of the four models considered when comparing RMSE $\times$ Execution Time. While some applications obviously require the most precise propagation possible, in certain cases it might be necessary to carefully balance RMSE and computational cost when selecting an appropriate model for Orbit Propagation, and the models presented in this work are an interesting solution.

\section{Conclusion and Future Work}

We presented a Machine Learning framework for Orbit Propagation, using data from the LARETS satellite. Furthermore, akin to numerical methods, the model is given features that represent forces that affect the object's orbit. The results are yet to achieve the same precision of an accurate numerical propagator with a full force model, however the results are promising, specially considering the remarkable reduction in computation time achieved by the iTransformer + FNN model. A somewhat surprising result is that the simpler model, in terms of complexity of the coarse model, achieved better performance, which might be due to the robustness of Prophet at forecasting unseen data. An elements that the authors would like to point out, is that a forecasting scheme of one-shot forecasting seems more suited to this task, as it can propagate further without much error growth, eventually becoming more precise than a J2 propagator.

In future work, the authors propose extending the exogenous variables added to the model, including a Solar Radiation Pressure model. An element that was not used in this work is employing exogenous features known in the future, such as for example, an estimation of the termospheric density during the forecasting period being first calculated and then used as an added variable. Since this replicates the procedure of the numerical propagator, it seems like a viable option to further improve ML models.
Furthermore, while keplerian and equinoctial parameters were added to the extended input, there might be value in using other coordiantes, specifically equinoctial elements, as forecasting targets. 

A step forward in this research line will necessarily require the expansion of the ML model to more than 1 satellite, as any useful propagation model requires some level of generalizability  to unseen space objects, which will in turn require the introduction of non-dynamic features, such as shape, mass, reflectivity, etc., to the model input.


\section*{Acknowledgment}

This work was partially supported by FCT/PT Space under the PhD grant PRT/BD/153601/2021 and the strategic project NOVA LINCS (UIDB/ 04516/2020) with the financial support of Project “Artificial Intelligence Fights Space Debris” (N C626449889-0046305) co-funded by Recovery and Resilience Plan and NextGeneration EU Funds, www.recuperarportugal.gov.pt. Funded by the European Union Project (TARDIS, 101093006). Views and opinions expressed are however those of the author(s) only and do not necessarily reflect those of the European Union. Neither the European Union nor the granting authority can be held responsible for them.

\noindent\includegraphics[width=\columnwidth]{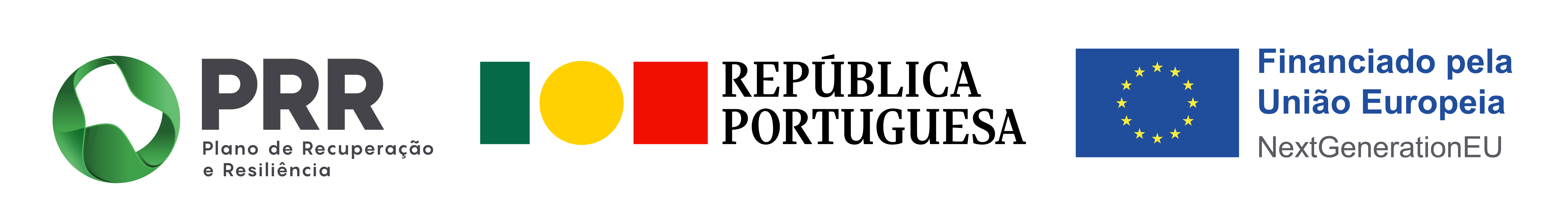}

\bibliographystyle{IEEEtran}
\bibliography{references}

\end{document}